\documentclass[letterpaper,10pt,conference]{ieeeconf}  

\IEEEoverridecommandlockouts                              

\usepackage{graphics,graphicx} 
\usepackage{epsfig} 
\usepackage{amsmath} 
\usepackage{amssymb}  
\usepackage{enumerate}
\usepackage{flushend}
\usepackage{cite}
\usepackage{color}
\usepackage{float}
\usepackage{wasysym}
\usepackage{eso-pic}
\def\bEta{\mbox{\boldmath$\eta$}}
\def\bAs{\mbox{\scriptsize{\boldmath$A$}}}

\def\dth{\mbox{$\dot{\theta}$}}

\def\bb{\mbox{\boldmath$b$}}
\def\bq{\mbox{\boldmath$q$}}

\def\bg{\mbox{\boldmath$g$}}
\def\bJ{\mbox{\boldmath$J$}}

\def\bM{\mbox{\boldmath$M$}}

\def\bv{\mbox{\boldmath$v$}}

\def\bb{\mbox{\boldmath$b$}}

\def\bV{\mbox{\boldmath$V$}}
\def\bA{\mbox{\boldmath$A$}}
\def\bI{\mbox{\boldmath$I$}}
\def\bG{\mbox{\boldmath$G$}}

\def\bx{\mbox{\boldmath$x$}}

\def\by{\mbox{\boldmath$y$}}

\def\bu{\mbox{\boldmath$u$}}

\def\blambda{\mbox{\boldmath$\lambda$}}

\def\bS{\mbox{\boldmath$S$}}

\def\bC{\mbox{\boldmath$C$}}

\def\mR{\mathbb{R}}

\title{\LARGE \bf
Asymptotically Stable Gait Generation and Instantaneous Walkability Determination for Planar Almost Linear Biped with Knees}

\author{Fumihiko Asano, Ning Lei, and Taiki Sedoguchi
\thanks{This research was partially supported by Grant-in-Aid for Scientific Research (C) No. 23K03727, provided by the Japan Society for the Promotion of Science (JSPS).}
\thanks{The authors are with the Graduate School of Advanced Science and Technology, Japan Advanced Institute of Science and Technology, 1-1 Asahidai, Nomi, Ishikawa 923-1292, Japan {\tt\small \{fasano,s2510193,sedoguchi\}@jaist.ac.jp}}}

\AddToShipoutPictureFG*{
\AtPageLowerLeft{
\ifnum\value{page}=1
\put(10,10){
\parbox{0.9\paperwidth}{
\tiny
\textcopyright\ 2026 IEEE. This is the author's version of the work.
A typographical error in the inertia matrix $\bM$ has been corrected.
}}
\fi
}
}

\begin{document}

\maketitle
\thispagestyle{empty}
\pagestyle{empty}

\setlength{\arraycolsep}{1.5pt}
\setlength{\abovedisplayskip}{5.6pt}
\setlength{\belowdisplayskip}{5.6pt}
\begin{abstract}
A class of planar bipedal robots with unique mechanical properties has been proposed, where all links are balanced around the hip joint, preventing natural swinging motion due to gravity. A common property of their equations of motion is that the inertia matrix is a constant matrix, there are no nonlinear velocity terms, and the gravity term contains simple nonlinear terms. By performing a Taylor expansion of the gravity term and making a linear approximation, it is easy to derive a linearized model, and calculations for future states or walkability determination can be performed instantaneously without the need for numerical integration. This paper extends the method to a planar biped robot model with knees. First, we derive the equations of motion, constraint conditions, and inelastic collisions for a planar 6-DOF biped robot, design its control system, and numerically generate a stable bipedal gait on a horizontal plane. Next, we reduce the equations of motion to a 3-DOF model, and derive a linearized model by approximating the gravity term as linear around the expansion point for the thigh frame angle. Through numerical simulations, we demonstrate that calculations for future states and walkability determination can be completed in negligible time. By applying control inputs to the obtained model, performing state-space realization, and then discretizing it, instantaneous walkability determination through iterative calculation becomes possible. Through detailed gait analysis, we discuss how the knee joint flexion angle and the expansion point affect the accuracy of the linear approximation, and the issues that arise when descending a small step.
\end{abstract}
\section{INTRODUCTION}

The primary reason why the efficient gait generation theory for legged locomotion robots, originating from passive dynamic walking research \cite{McGeer,McGeer2}, has yet to permeate industry is the difficulty of having to spend an enormous amount of time analyzing whether it can safely walk to the target destination. While recent advances in visual information processing enable high-precision, high-speed recognition of road surface conditions, determining the stability of hybrid zero dynamics (HZD) \cite{Grizzle,HZD} remains complex, and theoretical improvements for practical implementation have not progressed. To safely operate bipedal robots in complex environments, it is essential to determine in real time not only whether the inherently generated gait is stable but also whether safe walking to the target destination is feasible.

The objective of our study is to generate asymptotically stable gaits for a planar biped robot described by an almost linear equation of motion, and to construct an ultrahigh-speed future state calculation method that does not require numerical integration by utilizing the approximate HZD. Compared to general bipedal humanoids, the biped robots addressed in this study have a slightly different appearance, but they possess the mathematical characteristics described below. Therefore, even when performing approximate linearization of motion, significant differences between the original nonlinear model and the linearized model are unlikely to arise, enabling high-precision future state calculations without the need for numerical integration.

Kieffer and Bale proposed an almost linear biped consisting of two leg frames and a reaction wheel sandwiched between them, and examined the wheel gait generation problem \cite{Kiefer}. The mathematical characteristic of the robot is that the centers of mass (COMs) of the leg frames and the reaction wheel are located at the same position as the hip joint that connects them, which results in a very simple equation of motion. Generating stable wheel gait is not an easy problem, but since the swing leg rotates in the same direction as the stance leg, there is no risk of swing-foot scuffing at mid-stance. Subsequently, Spong et al. examined feedback linearization for gait generation using the same model \cite{Spong}. Furthermore, Agrawal and Fattah proposed a planar biped model with knees and the COMs of both legs at the same position as the hip joint, and discussed the gait generation problem for one step \cite{Agrawal}. A common feature of these almost linear legged locomotion (AL3) robots is that their equations of motion have a constant inertia matrix, no nonlinear velocity term, and nonlinear terms in the gravity term that can be easily approximated as linear.

This paper reconstructs the biped model proposed in \cite{Agrawal,Agrawal2} and develops an instantaneous walkability determination method through its dimensionality reduction, approximate linearization, and discretization. Numerical simulations analyze how the knee bending angle and the expansion point of the thigh frame angle in the approximate linearization affect the approximation accuracy. Furthermore, considering the situation of descending a small step during walking, we discuss issues in the control system that require improvement through this instantaneous walkability determination.


\section{NONLINEAR MODEL AND STABLE GAIT GENERATION}

\subsection{Equations of Motion and Constraint Conditions}

\begin{figure*}[!t]
\centering
\includegraphics[width=0.95\linewidth]{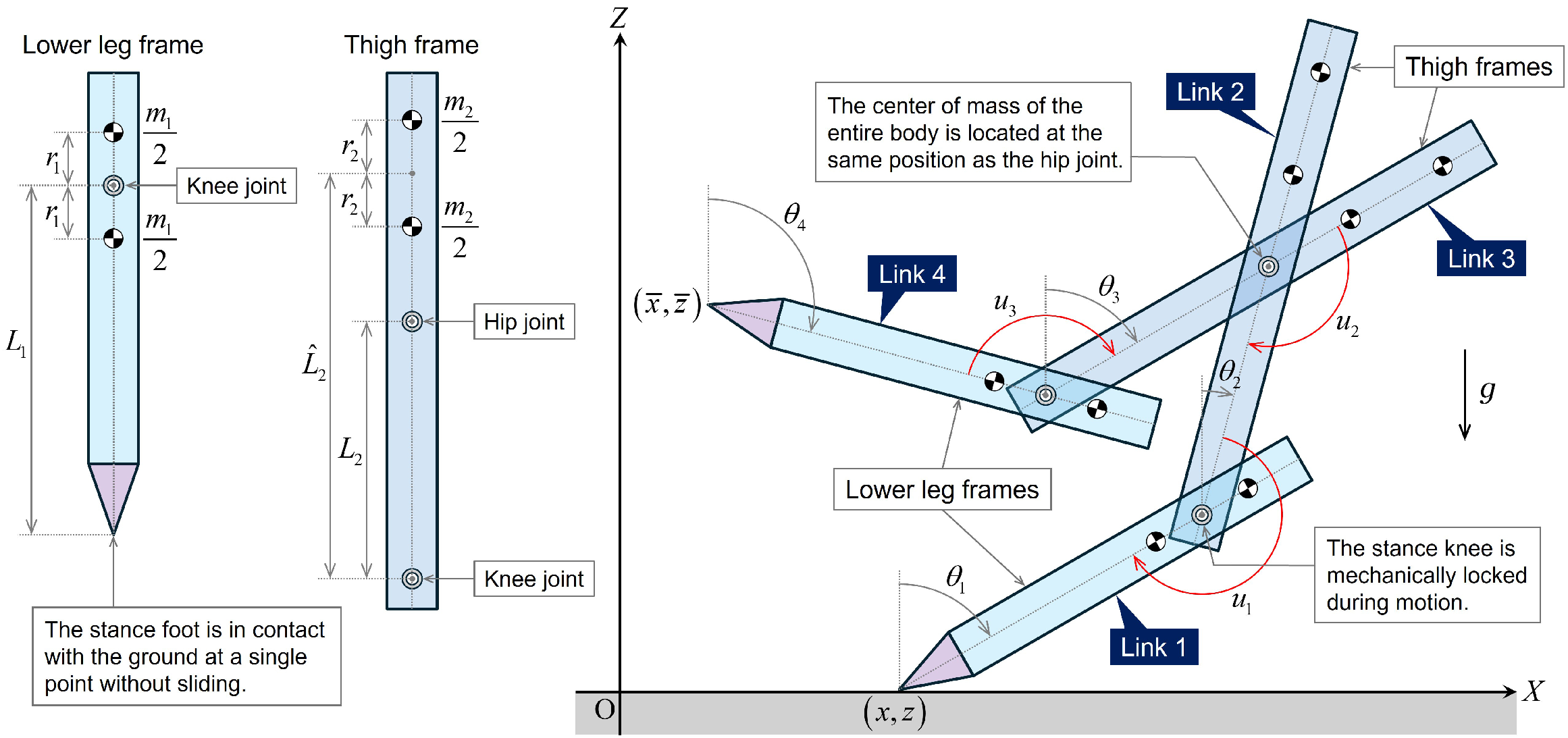}\\
\vspace*{-2mm}
\caption{Model of planar, 6-DOF, almost linear biped with knees}
\label{fig2.01}
\vspace*{-3mm}
\end{figure*}

Figure \ref{fig2.01} shows the model of the planar 6-DOF biped robot with knees discussed in this paper. The position of the stance foot is $(x,z)$, the lower leg of the stance leg is Link 1, the thigh of it is Link 2, the thigh of the swing leg is Link 3, the lower leg of it is Link 4, and the absolute angle of Link $k$ from the vertical upward direction is $\theta_k$. The stance foot is assumed to be in contact with the ground at a single point and not to slide during motion. Additionally, rotational torques $u_1$, $u_2$, and $u_3$ can be applied to the knee joint of the stance leg, the hip joint, and the knee joint of the swing leg, respectively. The stance knee is, however, mechanically locked during motion, and $u_1$ is effectively non-functional.

Let the masses of the lower leg and thigh frames be $m_1$ and $m_2$, respectively. On each frame, the respective masses are divided equally and placed at a distance $r_1$ and $r_2$ from the COM. As a result, the moments of inertia around the COM are $I_1 = m_1 r_1^2$ and $I_2 = m_2 r_2^2$, respectively. As shown in Fig. \ref{fig2.01} left, let $L_1$ denote the length from the lower end of the lower leg frame to the knee joint (equivalent to the COM), $L_2$ denote the length from the knee joint to the hip joint on the thigh frame, and $\hat{L}_2$ denote the length from the knee joint to the COM. The condition for the COM of the thigh frame to align with the hip joint when the lower leg frame is attached becomes
\begin{equation}
\hat{L}_2 = L_2 \left( 1 + \frac{m_1}{m_2} \right).
\end{equation}
Since the COM of the lower leg is at the same position as the knee joint, no natural swinging motion due to gravity occurs around this point. The COM of the upper leg is above the hip joint, and the masses of the upper and lower legs balance each other, resulting in the overall COM of the leg being at the same position as the hip joint. In other words, around the hip joint, none of the links exhibit natural swinging motion due to gravity.

Let $\bq = 
\left[
\begin{array}{cccccc}
x & z & \theta_1 & \theta_2 & \theta_3 & \theta_4
\end{array}
\right]^{\rm T}
$ be the generalized coordinate vector. The robot equation of motion then becomes
\begin{equation}
\bM \ddot{\bq} + \bC \dot{\bq} + \bg = \bS \bu + \bJ_c^{\rm T} \blambda_c,
\label{eq2.01}
\end{equation}
where
\setlength{\arraycolsep}{0.8pt}
{\small
\begin{eqnarray*}
\bM \! &=& \!
\left[
\begin{array}{cccccc}
m & 0 & m L_1 C_1 & m L_2 C_2 & 0 & 0 \\
 & m & -m L_1 S_1 & -m L_2 S_2 & 0 & 0 \\
 & & m L_1^2 + I_1 & m L_1 L_2 C_{12} & 0 & 0 \\
 & & & \frac{(m_1+2m_2) m L_2^2}{2m_2} + I_2 & 0 & 0 \\
 & & & & \frac{m_1 m L_2^2}{2 m_2} + I_2 & 0 \\
{\rm Sym.} & & & & & I_1
\end{array}
\right],
\\
\bC \! &=& \!
\left[
\begin{array}{cccccc}
0 & 0 & -m L_1 \dth_1 S_1 & -m L_2 \dth_2 S_2 & 0 & 0 \\
0 & 0 & -m L_1 \dth_1 C_1 & -m L_2 \dth_2 C_2 & 0 & 0 \\
0 & 0 & 0 & m L_1 L_2 \dth_2 S_{12} & 0 & 0 \\
0 & 0 & -m L_1 L_2 \dth_1 S_{12} & 0 & 0 & 0 \\
0 & 0 & 0 & 0 & 0 & 0 \\
0 & 0 & 0 & 0 & 0 & 0
\end{array}
\right] \hspace*{-0.5mm},
\bg \! = \! 
\left[
\begin{array}{c}
0 \\
m g \\
-m g L_1 S_1 \\
-m g L_2 S_2 \\
0 \\
0
\end{array}
\right].
\label{eq2.02}
\end{eqnarray*}
}
\setlength{\arraycolsep}{1.5pt}
Here, $m := 2 (m_1 + m_2)$ is the robot's total mass, $C_1 := \cos \theta_1$, $S_1 := \sin \theta_1$, $C_2 := \cos \theta_2$, $S_2 := \sin \theta_2$, $C_{12} := \cos (\theta_1 - \theta_2)$ and $S_{12} := \sin (\theta_1 - \theta_2)$.

The details of the driving matrix and control input vector are as follows.
\begin{equation}
\bS \bu = 
\small
\left[
\begin{array}{ccc}
0 & 0 & 0 \\
0 & 0 & 0 \\
1 & 0 & 0 \\
-1 & 1 & 0 \\
0 & -1 & 1 \\
0 & 0 & -1
\end{array}
\right] \!\!
\left[
\begin{array}{c}
u_1 \\
u_2 \\
u_3
\end{array}
\right] 
\label{eq2.025}
\end{equation}
As mentioned, however, the stance knee is mechanically locked during motion, and $u_1$ is effectively non-functional. Therefore, only $u_2$ and $u_3$ need to be designed.

The velocity constraint conditions that the stance foot is in contact with the ground at a single point without slipping are described $\dot{x} = 0$ and $\dot{z} = 0$. The relative angle between the lower leg link and upper leg link of the stance leg, i.e., the bending angle of the stance knee, is assumed to be mechanically fixed at $\beta$ at all times. This geometric condition is described as $\theta_1 - \theta_2 = \beta$, and by differentiating this with respect to time, we obtain the velocity constraint condition $\dth_1 - \dth_2 = 0$. Summarizing these three equations yields
\begin{equation}
\bJ_c \dot{\bq} = 
\left[
\begin{array}{cccccc}
1 & 0 & 0 & 0 & 0 & 0 \\
0 & 1 & 0 & 0 & 0 & 0 \\
0 & 0 & 1 & -1 & 0 & 0
\end{array}
\right] \dot{\bq} = {\bf 0}_{3 \times 1}.
\label{eq2.03}
\end{equation}
The time derivative of this becomes $\bJ_c \ddot{\bq} = {\bf 0}_{3 \times 1}$. Using this and Eq. (\ref{eq2.01}), the undetermined multiplier vector $\blambda_c \in {\mR}^3$ can be determined. Its first component is the horizontal ground reaction force $F_x$, the second component is the vertical ground reaction force $F_z$, and the third component is the constraint force at the stance knee.

One condition necessary for stable gait generation is that $F_z$ remains positive at all times. The model in Fig. \ref{fig2.01} maintains the entire COM at the same position as the hip joint throughout. Since the COM moves as a single inverted pendulum, even when the hip joint or swing-knee joint undergoes high-speed rotational motion, the COM does not fluctuate violently up and down, and $F_z$ remains stable.

\subsection{Collision Equation}

As shown in Fig. \ref{fig2.01} right, if the position of the swing foot is denoted as $(\bar{x},\bar{z})$, the landing occurs at the instant when $\bar{z}$ decreases monotonically while maintaining a positive value and reaches zero. $\bar{z}$ is specifically described by
\begin{equation}
\bar{z} = z + L_1 \cos \theta_1 + L_2 \cos \theta_2 - L_2 \cos \theta_3 - L_1 \cos \theta_4.
\label{eq2.04}
\end{equation}
After this completely inelastic collision, the swing foot or fore foot is assumed to contact the ground at a single point and not slide. The inelastic collision equation becomes
\begin{equation}
\bM \dot{\bq}^+ = \bM \dot{\bq}^- + \bJ_I^{\rm T} \blambda_I.
\label{eq2.05}
\end{equation}
The second term on the right-hand side of this equation is the impulse vector, and its Jacobian $\bJ_I \in \mR^{4 \times 6}$ is determined as follows. The velocity constraint conditions, which state that the swing foot collides inelastically with the ground and remains in contact without sliding immediately after the collision, are specified as follows.
\begin{eqnarray}
\!\!\!
\dot{\bar{x}}^+ &=& \tfrac{\rm d}{{\rm d}t} \left(
x + L_1 S_1 + L_2 S_2 - L_2 S_3 - L_1 S_4
\right)^+ = 0 \\
\!\!\!
\dot{\bar{z}}^+ &=& \tfrac{\rm d}{{\rm d}t} \left(
z + L_1 C_1 + L_2 C_2 - L_2 C_3 - L_1 C_4
\right)^+ = 0
\label{eq2.06}
\end{eqnarray}
In addition to these, the velocity constraint conditions that the knee joints of both legs are mechanically locked at the same angle $\beta$ during the collision phase are specified as $\dth_1^+ - \dth_2^+ = 0$ and $\dth_3^+ - \dth_4^+ = 0$. In other words, the robot causes inelastic collision equivalent to that of a compass-like biped robot with two bent legs \cite{AIM2015}. Summarizing these four conditions yields
\begin{equation}
\bJ_I \dot{\bq}^+ \! = \!
\left[
\begin{array}{cccccc}
1 & 0 & L_1 C_1 & L_2 C_2 & -L_2 C_3 & -L_1 C_4 \\
0 & 1 & -L_1 S_1 & -L_2 S_2 & L_2 S_3 & L_1 S_4 \\
0 & 0 & 1 & -1 & 0 & 0 \\
0 & 0 & 0 & 0 & 1 & -1
\end{array}
\right] \dot{\bq}^+ \! = \! {\bf 0}_{4 \times 1}.
\label{eq2.07}
\end{equation}
At this stage, note that the angular positions included in $\bJ_I$ is immediately before impact, as the exchange between the stance leg and swing leg has not yet been considered. Solving for $\dot{\bq}^+$ from the simultaneous equations (\ref{eq2.05}) and (\ref{eq2.07}), and further considering the exchange of coordinates between the stance and swing legs (rear and fore legs), the velocity vector immediately after the collision can be obtained as
\begin{equation}
\dot{\bq}^+ = 
\left[
\begin{array}{cccccc}
0 &
0 &
\xi \dth_1^- &
\xi \dth_1^- &
\dth_1^- &
\dth_1^-
\end{array}
\right]^{\rm T},
\label{eq2.08}
\end{equation}
where
\begin{eqnarray}
\xi &:=& \frac{N_1}{D_1}, \\
N_1 &=& m_1 (m_1 + m_2) L_2^2 + m_2 (I_1 + I_2) \nonumber \\
 & & + m_2 m \cos \alpha \left( L_1^2 + L_2^2 + 2 L_1 L_2 \cos \beta \right), \\
D_1 &=& (m_1 + m_2) (m_1 + 2 m_2) L_2^2 + m_2 \left( m L_1^2 + I_1 + I_2 \right) \nonumber \\
 & & + 2 m_2 m L_1 L_2 \cos \beta.
\label{eq2.09}
\end{eqnarray}

\subsection{Control System Design}

This subsection provides only an overview of output following control. The robot has three active joints, and the relative joint angle controlled by the control torque $u_k$ generates motion according to the acceleration command signal $v_k$. In this paper, however, since the stance knee is mechanically locked, the two controllable active joints are the hip and swing knee joints. Their relative angles are chosen as the control output vector as follows.
\begin{equation}
\by :=
\left[
\begin{array}{c}
\theta_2 - \theta_3 \\
\theta_3 - \theta_4
\end{array}
\right] 
\label{eq2.10}
\end{equation}
Hereafter, let $t$ be a time variable reset to zero at each collision, and $T_{\rm set}$ be the target settling time. The relative hip-joint angle is $-\alpha$ immediately after impact, that is, at $t = 0^+$, and we interpolate it using a fifth-order time function so that it becomes $\alpha$ at $t = T_{\rm set}$. The relative joint angle of the swing knee is $-\beta$ immediately after impact, and we bend it further by $\gamma$ from there, then follow a time function of the cube of sin so that it becomes $-\beta$ again at $t = T_{\rm set}$. We then determine the target time trajectory vector as follows.
\begin{equation}
\by_{\rm d} (t) := \left\{
\begin{array}{ll}
\left[
\begin{array}{c}
\sum_{k=0}^{5} a_k t^k \\
- \beta - \gamma \sin^3 
\left(
\frac{\pi t}{T_{\rm set}}
\right)
\end{array}
\right] 
 & \left( 0 \leq t \leq T_{\rm set} \right) \\
\left[
\begin{array}{c}
\alpha \\
- \beta
\end{array}
\right] 
 & \left( t > T_{\rm set} \right)
\end{array}
\right.
\label{eq2.11}
\end{equation}
We require that, however, the target impact posture must be achieved at $t = T_{\rm set}$ before the next impact as a necessary condition for stable gait generation. Immediately before each impact, the robot mechanically locks both knee joints and falls down as a 1-DOF rigid body.

By designing and applying control inputs that achieve $\ddot{\by} = \bv$ using $\bv = \ddot{\by}_{\rm d} (t) \in \mR^{2}$ as the acceleration command vector, we can generate an asymptotically stable bipedal gait on the horizontal plane as described below. To avoid including PD feedback terms in $\bv$, the coefficients of the target fifth-order time function are designed so that not only the hip joint angle but also the angular velocity immediately after impact match the actual values. The specific coefficients are obtained as
\begin{eqnarray*}
a_5 \! &=& \! \frac{12 \alpha - 3 (\xi-1) \dth_1^- T_{\rm set}}{T_{\rm set}^5}, \ 
a_4 \! = \! \frac{-30 \alpha + 8 (\xi-1) \dth_1^- T_{\rm set}}{T_{\rm set}^4}, \\
a_3 \! &=& \! \frac{20 \alpha - 6 (\xi-1) \dth_1^- T_{\rm set}}{T_{\rm set}^3}, \ 
a_1 \! = \! (\xi - 1) \dth_1^-, \ a_0 \! = \! -\alpha,
\label{eq2.12}
\end{eqnarray*}
and $a_2 = 0$. Note that, however, since these coefficients include $\dth_1^-$, recalculation is required for each collision.

The method for determining the control input will be detailed in the next section when discussing control system design for the linearized reduced model; therefore, specifics are omitted here.

\begin{figure}[!b]
\vspace*{-3mm}
\centering
\footnotesize
\scalebox{0.65}{
\input{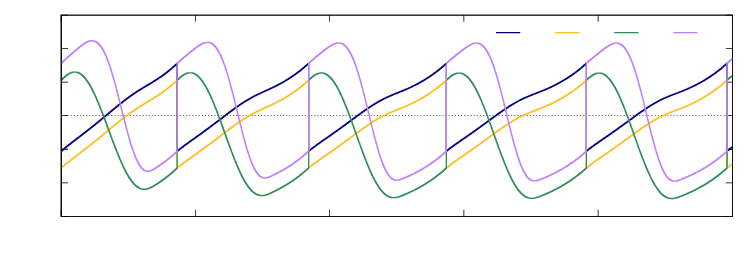}
}\\
\vspace*{-1mm}
(a) Angular positions\\
\scalebox{0.65}{
\input{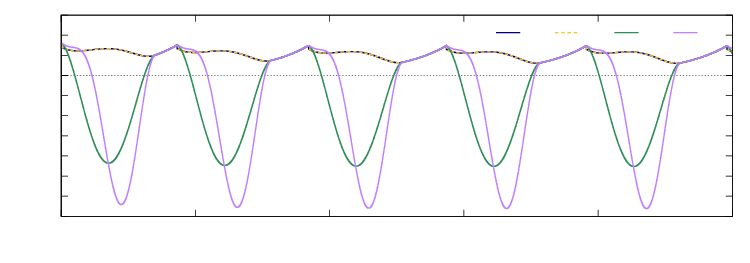}
}\\
\vspace*{-1mm}
(b) Angular velocities\\
\scalebox{0.65}{
\input{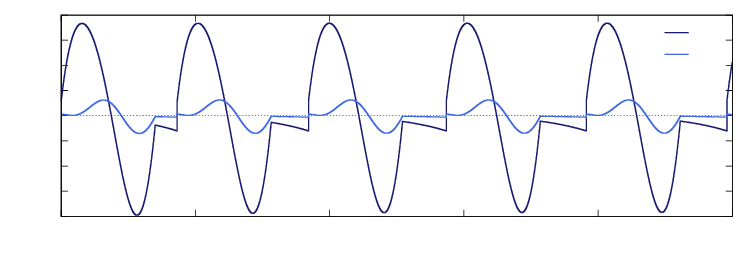}
}\\
\vspace*{-1mm}
(c) Control inputs\\
\scalebox{0.65}{
\input{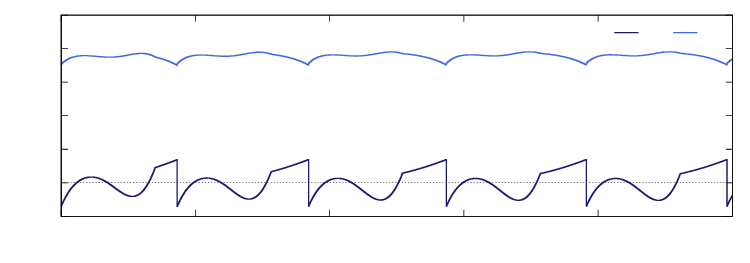}
}\\
\vspace*{-1mm}
(d) Ground reaction forces\\
\scalebox{0.65}{
\input{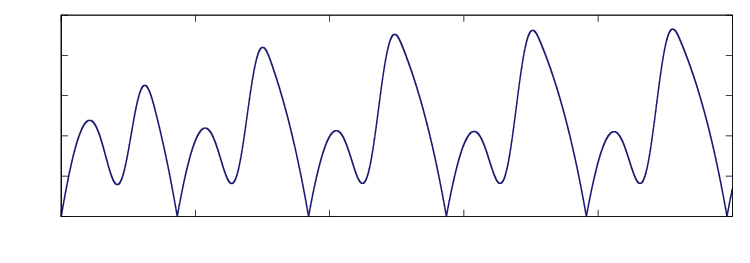}
}\\
\vspace*{-1mm}
(e) $Z$-position of swing foot
\vspace*{-2mm}
\caption{Simulation results of asymptotically stable walking of kneed biped}
\label{fig2.02}
\end{figure}

\subsection{Typical Bipedal Gait}

Figure \ref{fig2.02} shows the simulation results of generating an asymptotically stable gait on a horizontal plane. Here, (a) is the angular positions, (b) the angular velocities, (c) the control inputs, (d) the ground reaction forces, and (e) the $Z$-position of the swing foot. The robot's physical and control parameters were chosen as the values listed in Table \ref{table01}. The robot started walking from the target impact posture where the velocity vector was set as in Eq. (\ref{eq2.08}) with $\dth_1^- = 0.8$ [rad/s]. From Fig. \ref{fig2.02}(b), we can see that during motion, $\dth_1 = \dth_2$ always holds true. Furthermore, after $t = T_{\rm set}$, all angular velocities change while maintaining the same value. In the reduced-dimensional model discussed in the subsequent sections, the lower-leg angle $\theta_1$ will no longer appear in the equations. When referring to the angular velocity immediately before impact, however, we will consistently use the notation $\dth_1^-$. Fig. \ref{fig2.02}(d) shows that the vertical ground reaction force $F_z$ exhibits a stable waveform with minimal variation. As mentioned, this is because the entire COM performs simple rotational motion as a single inverted pendulum. This is also a common characteristic of the AL3 robot. From Fig. \ref{fig2.02}(e), we can see that the swing-foot clearance is always maintained, enabling walking without scuffing the floor.

\begin{table}[!t]
\vspace*{1.8mm}
\caption{Physical and control parameters}
\label{table01}
\vspace*{-2mm}
\centering
\renewcommand{\arraystretch}{1.4}
\begin{tabular}{ccc} \hline
$m_1$ & 1.0 & kg \\
$m_2$ & 1.0 & kg \\
$L_1$ & 0.5 & m \\
$L_2$ & 0.5 & m \\
$r_1$ & 0.25 & m \\
$r_2$ & 0.25 & m \\ \hline
\end{tabular}
\hspace*{1mm}
\begin{tabular}{ccc} \hline
$I_1 \ \left( =m_1 r_1^2 \right)$ & 0.0625 & kg$\cdot$m$^2$ \\
$I_2 \ \left( =m_2 r_2^2 \right)$ & 0.0625 & kg$\cdot$m$^2$ \\ 
$\alpha$ & $\frac{\pi}{6}$ & rad \\
$\beta$ & 0.1 & rad \\
$\gamma$ & 0.3 & rad \\
$T_{\rm set}$ & 0.7 & s \\ \hline
\end{tabular}
\vspace*{-3mm}
\end{table}

\section{REDUCED LINEARIZED MODEL AND DISCRETIZATION}

\subsection{Linearized Reduced Model}

This section describes a method for instantaneously calculating future states through iterative computation by reducing, linearizing, and discretizing the nonlinear model.

Considering that the velocity constraint condition in Eq. (\ref{eq2.03}) always holds, the velocity vector can be reduced to three dimensions as follows.
\begin{equation}
\dot{\bq} = 
{\small
\left[
\begin{array}{ccc}
0 & 0 & 0 \\
0 & 0 & 0 \\
1 & 0 & 0 \\
1 & 0 & 0 \\
0 & 1 & 0 \\
0 & 0 & 1
\end{array}
\right] \!\!
\left[
\begin{array}{c}
\dth_2 \\
\dth_3 \\
\dth_4
\end{array}
\right]
} =: \bV \dot{\bar{\bq}}
\label{eq3.01}
\end{equation}
Using this relationship, multiplying both sides of Eq. (\ref{eq2.01}) by $\bV^{\rm T}$ from the left yields the following 3-DOF model.
\begin{equation}
\bar{\bM} \ddot{\bar{\bq}} + \bar{\bg} = \bar{\bS} \bar{\bu}
\label{eq3.02}
\end{equation}
The details of each term are as follows.
\begin{eqnarray*}
\bar{\bM} &=&
\left[
\begin{array}{ccc}
\bar{M}_{11} & 0 & 0 \\
0 & \frac{m_1 m L_2^2}{2m_2} + I_2 & 0 \\
0 & 0 & I_1
\end{array}
\right], \ \ \bar{\bS} \bar{\bu} = 
\left[
\begin{array}{cc}
1 & 0 \\
-1 & 1 \\
0 & -1
\end{array}
\right] 
\hspace*{-1mm}
\left[
\begin{array}{c}
u_2 \\
u_3
\end{array}
\right] \\
\bar{\bg} &=&
\left[
\begin{array}{c}
-mg \left( L_1 \sin \left( \theta_2 + \beta \right) + L_2 S_2 \right) \\
0 \\
0
\end{array}
\right], \ \ \bar{\bq} = 
\left[
\begin{array}{c}
\theta_2 \\
\theta_3 \\
\theta_4
\end{array}
\right] \\
\bar{M}_{11} &=& m L_1^2 + \frac{(m_1 + 2 m_2)m L_2^2}{2m_2} + 2 m L_1 L_2 \cos \beta \nonumber \\
 & & + I_1 + I_2
\label{eq3.03}
\end{eqnarray*}
Defining the first component of the $\bar{\bg}$ vector as $\bar{g}_1 \left( \theta_2 , \beta \right)$ and its partial derivative with respect to $\theta_2$ as $\bar{g}' \left( \theta_2 , \beta \right)$, the linear approximation of this vector around $\theta_2 = \theta_2^{\ast}$ then yields
\begin{eqnarray*}
\bar{\bg} \! &\approx& \!
\left[
\begin{array}{ccc}
\bar{g}'_{1} \left( \theta_2^{\ast} , \beta \right)
 & 0 & 0 \\
0 & 0 & 0 \\
0 & 0 & 0
\end{array}
\right]
 \!\!
\left[
\begin{array}{c}
\theta_2 \\
\theta_3 \\
\theta_4
\end{array}
\right] + 
\left[
\begin{array}{c}
\bar{g}_1 \left( \theta_2^{\ast} , \beta \right) - \bar{g}'_{1} \left( \theta_2^{\ast} , \beta \right) \theta_2^{\ast} \\
0 \\
0 \\
\end{array}
\right] \\
&=:& \bar{\bG} \bar{\bq} + \bar{\bg}_{\beta}.
\label{eq3.04}
\end{eqnarray*}
We finally obtain the reduced linearized model as
\begin{equation}
\bar{\bM} \ddot{\bar{\bq}} + \bar{\bG} \bar{\bq} + \bar{\bg}_{\beta} = \bar{\bS} \bar{\bu}.
\label{eq3.05}
\end{equation}
Note that the $\bar{\bg}_{\beta}$ vector acts as a propulsive force in the walking direction. For example, when $\theta_2^{\ast} = 0$, the first component of this vector becomes $
\bar{g}_1 \left( 0 , \beta \right) = -mgL_1 \sin \beta$, and this is the same mechanical effect as applying a control torque $mgL_1 \sin \beta$ around the ground contact point of the stance leg. When $\beta$ is maintained at zero, that is, when walking with the stance knee fully extended, this propulsive force vanishes.

\subsection{Discretization of Controlled Reduced Linearized Model}

Hereinafter, we assume that the robot starts walking from the target impact posture immediately after impact; this is defined as the 0-th impact. The next fore-foot impact is the 1-st impact, and the motion between the 0-th and 1-st impacts is called the ``0-th step''. The subsequent impacts and steps are contextually counted. The period from the $i$-th impact to the $i+1$-th impact is defined as the $i$-th step period, $T_i$. The steady step period is then written as $T_{\infty}$.

The second-order derivative of $\by$ in Eq. (\ref{eq2.10}) with respect to time becomes
\begin{equation}
\ddot{\by} = \bar{\bS}^{\rm T} \ddot{\bar{\bq}} = 
\bar{\bS}^{\rm T} \bar{\bM}^{-1} \left(
\bar{\bS} \bar{\bu} - \bar{\bG} \bar{\bq} - \bar{\bg}_{\beta}
\right).
\label{eq3.06}
\end{equation}
Then, we can determine the control input $\bar{\bu}$ for achieving $\ddot{\by} = \bv$ as
\begin{equation}
\bar{\bu} = \left(
\bar{\bS}^{\rm T} \bar{\bM}^{-1} \bar{\bS} 
\right)^{-1}
\left(
\bv + \bar{\bS}^{\rm T} \bar{\bM}^{-1} \left( \bar{\bG} \bar{\bq} + \bar{\bg}_{\beta}
\right)
\right).
\label{eq3.07}
\end{equation}
By substituting this into Eq. (\ref{eq3.04}) and arranging, the state-space realization of the controlled linearized reduced (CLRed) model becomes
\begin{equation}
\dot{\bx} = \bA \bx + \bb_1 + \bb_2 v_2 + \bb_3 v_3,
\label{eq3.08}
\end{equation}
where
{\small
\begin{eqnarray*}
\bA &=& \left[
\begin{array}{cc}
{\bf 0}_{3 \times 3} & \bI_3 \\
\bar{\bM}^{-1} \left(
\bar{\bS} \left( \bar{\bS}^{\rm T} \bar{\bM}^{-1} \bar{\bS} \right)^{-1} \bar{\bS}^{\rm T} \bar{\bM}^{-1} - \bI_3
\right) \bar{\bG} & {\bf 0}_{3 \times 3}
\end{array}
\right], \\
\bb_1 &=& \left[
\begin{array}{c}
{\bf 0}_{3 \times 1} \\
\bar{\bM}^{-1} \left(
\bar{\bS} \left( \bar{\bS}^{\rm T} \bar{\bM}^{-1} \bar{\bS} \right)^{-1} \bar{\bS}^{\rm T} \bar{\bM}^{-1} - \bI_3
\right) \bar{\bg}_{\beta}
\end{array}
\right], \\
\bb_2 &=& \left[
\begin{array}{c}
{\bf 0}_{3 \times 1} \\
\bar{\bM}^{-1}
\bar{\bS} \left( \bar{\bS}^{\rm T} \bar{\bM}^{-1} \bar{\bS} \right)^{-1} 
\left[
\begin{array}{c}
1 \\
0
\end{array}
\right]
\end{array}
\right], \\
\bb_3 &=& \left[
\begin{array}{c}
{\bf 0}_{3 \times 1} \\
\bar{\bM}^{-1}
\bar{\bS} \left( \bar{\bS}^{\rm T} \bar{\bM}^{-1} \bar{\bS} \right)^{-1} 
\left[
\begin{array}{c}
0 \\
1
\end{array}
\right]
\end{array}
\right].
\label{eq3.09}
\end{eqnarray*}
}
By solving Eq. (\ref{eq3.08}), the state at $t = T_{\rm set}$ in the $i$-th step can be obtained as
\begin{equation}
\bx \left( T_{\rm set} \right) = {\rm e}^{\bAs T_{\rm set}} \left(
\bx_i^+ + \bEta_1 + \bEta_{2i} + \bEta_3
\right),
\label{eq3.10}
\end{equation}
where
\begin{equation*}
\bEta_1 \! := \!\! \int_{0^+}^{T_{\rm set}} \hspace*{-3mm} {\rm e}^{- \bAs \tau} \bb_1 {\rm d}\tau, \ 
\bEta_k \! := \!\! \int_{0^+}^{T_{\rm set}} \hspace*{-3mm} {\rm e}^{- \bAs \tau} \bb_k v_k (\tau) {\rm d}\tau \ (k=2,3).
\label{eq3.11}
\end{equation*}
Note that $\bEta_2$ is defined as a definite integral calculation involving $v_2$, meaning it is a function vector containing the angular velocity immediately before the $i$-th impact, $\dth_{1i}^-$. Therefore, since $\bEta_2$ must be recalculated at each collision, it is denoted as $\bEta_{2i}$ in Eq. (\ref{eq3.10}).

After this, the robot falls forward as a 1-DOF rigid body, but in this phase, $v_2 = 0$ and $v_3 = 0$, so the state equation becomes $\dot{\bx} = \bA \bx + \bb_1$. This is redundant, however, and $\bA$ is not a regular matrix, so dimensionality reduction is performed for efficient calculation.
The $\bA$ matrix and $\bb_1$ vector have the following simple structure.
\begin{eqnarray*}
\bA &=& \! 
{\small
\left[
\begin{array}{cccccc}
0 & 0 & 0 & 1 & 0 & 0 \\
0 & 0 & 0 & 0 & 1 & 0 \\
0 & 0 & 0 & 0 & 0 & 1 \\
\omega^2 & 0 & 0 & 0 & 0 & 0 \\
\omega^2 & 0 & 0 & 0 & 0 & 0 \\
\omega^2 & 0 & 0 & 0 & 0 & 0 
\end{array}
\right]
} \!, \
\bb_1 = \! 
{\small
\left[
\begin{array}{c}
0 \\
0 \\
0 \\
b_1 \\
b_1 \\
b_1
\end{array}
\right]
} \!, \ \omega = \sqrt{\frac{N_2}{D_2}}, \ b_1 = \frac{N_3}{D_2} \\
N_2 &=& m_2 (m_1 + m_2) g \left( L_1 \cos \left( \theta_2^{\ast} + \beta \right) + L_2 \cos \theta_2^{\ast} \right) \\
N_3 &=& m_2 (m_1 + m_2) g \left( L_1 \sin \left( \theta_2^{\ast} + \beta \right) + L_2 \sin \theta_2^{\ast} \right) \\
 & & - m_2 (m_1 + m_2) g \theta_2^{\ast} \left( L_1 \cos \left( \theta_2^{\ast} + \beta \right) + L_2 \cos \theta_2^{\ast} \right) \\
D_2 &=& (m_1 + m_2)^2 L_2^2 + m_2 \left( (m_1 + m_2) L_1^2 + I_1 + I_2 \right) \\
 & & + m_2 m L_1 L_2 \cos \beta
\label{eq3.12}
\end{eqnarray*}

Extracting only the motion of $\theta_2$ yields the following two-dimensional state equation.
\begin{equation}
\frac{\rm d}{{\rm d}t}
\left[
\begin{array}{c}
\theta_2 \\
\dth_2
\end{array}
\right] = 
\left[
\begin{array}{cc}
0 & 1 \\
\omega^2 & 0
\end{array}
\right]
\left[
\begin{array}{c}
\theta_2 \\
\dth_2
\end{array}
\right] + 
\left[
\begin{array}{c}
0 \\
b_1
\end{array}
\right]
\label{eq3.13}
\end{equation}
Hereafter, this is denoted as $
\dot{\bar{\bx}} = \bar{\bA} \bar{\bx} + \bar{\bb}_1
$. Since the $\bar{\bA}$ matrix is regular, the state vector immediately before the next impact can be easily calculated as follows.
\begin{equation}
\bar{\bx}_{i+1}^- = {\rm e}^{\bar{\bAs} \left( T_i - T_{\rm set}\right)} 
\left(
\bar{\bx} \left( T_{\rm set} \right) + \bar{\bA}^{-1} \bar{\bb}_1
\right) - \bar{\bA}^{-1} \bar{\bb}_1
\label{eq3.14}
\end{equation}
In this phase, $\theta_2$ uniquely determines all other angles, so $\bar{z}$ can be calculated as follows.
\begin{eqnarray}
\bar{z} &=& L_1 \cos \left( \theta_2 + \beta \right) + L_2 \cos \theta_2 - L_2 \cos \left( \theta_2 - \alpha \right) \nonumber \\
& & - L_1 \cos \left( \theta_2 - \alpha + \beta \right)
\label{eq3.15}
\end{eqnarray}
The moment when this value decreases to zero can be determined numerically and instantaneously using the bisection method. Compared to numerical integration, the computational load is significantly lower.

\section{GAIT ANALYSIS}

\subsection{Comparison of Gait Descriptors in Stable Bipedal Gait}

\begin{figure}[!t]
\vspace*{0.5mm}
\centering
\footnotesize
\scalebox{0.65}{
\input{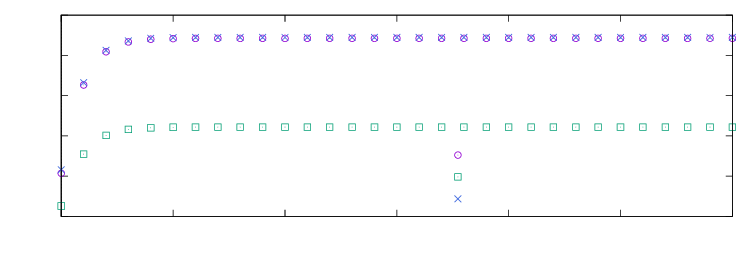}
}\\
\vspace*{-1mm}
(a) Step period \\
\scalebox{0.65}{
\input{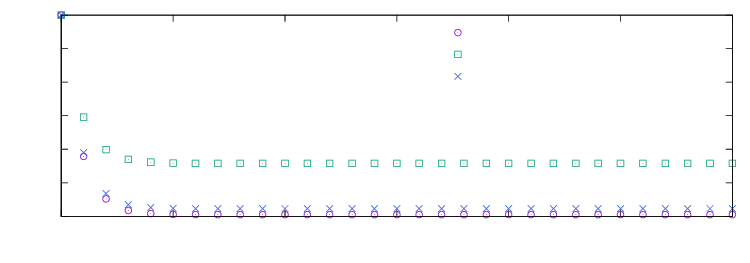}
}\\
\vspace*{-1mm}
(b) Angular velocity immediately before impact\\
\vspace*{-2mm}
\caption{Evolution of gait descriptors in asymptotically stable gaits of nonlinear and CLRed models where $\beta = 0.5$ [rad]}
\label{fig4.01}
\vspace*{-4mm}
\end{figure}

Figure \ref{fig4.01} plots the gait descriptors against the number of steps in the generated gaits of the nonlinear and CLRed models. Here, (a) is the step period, and (b) is the angular velocity immediately before impact. This time, $\beta$ was set to $0.5$ [rad], while all other system parameters were set to the same values as in Table \ref{table01}. The robot started walking from the target impact posture where the velocity vector was set as in Eq. (\ref{eq2.08}) with $\dth_1^- = 0.8$ [rad/s]. For the nonlinear model, the gait descriptors were calculated by conducting numerical integration; however, completing the calculations for 30 steps took several minutes. In the CLRed models, the gait descriptors were calculated instantaneously through iterative calculations, and the expansion points were set to two values: $\theta_{2}^{\ast} = 0$ and $\theta_{2}^{\ast} = -0.5 \beta = -0.25$ [rad]. The former approximates the gravity term as linear around the posture where the thigh frame faces vertically upward, while the latter does so around the posture where the entire body's COM or hip joint is directly above the stance foot.

From Fig. \ref{fig4.01}(a), it can be seen that the CLRed model with $\theta_2^{\ast} = -0.25$ [rad] shows step period values nearly identical to those of the nonlinear model, while the CLRed model with $\theta_2^{\ast} = 0$ [rad] shows significantly smaller values. From Fig. \ref{fig4.01}(b), it can be seen that the former shows slightly larger angular velocity values than the nonlinear model, while the latter shows significantly larger values. In the next subsection, we will analyze in greater detail how the approximation accuracy changes with respect to $\beta$ and $\theta_{2}^{\ast}$.

\subsection{Effects of $\beta$ and $\theta_2^{\ast}$ on Gait Characteristics and Approximation Accuracy}

Building on the results from the previous subsection, this subsection further generalizes the expansion point for gait analysis and discusses the approximation accuracy of the CLRed model. Treating $\kappa$ as a negative constant, the expansion point is defined by $\theta_2^{\ast} = \kappa \beta$.

\begin{figure}[!t]
\vspace*{0.5mm}
\begin{minipage}[t]{0.49\columnwidth}
\centering
\hspace*{-5mm}
\scalebox{0.46}{
\input{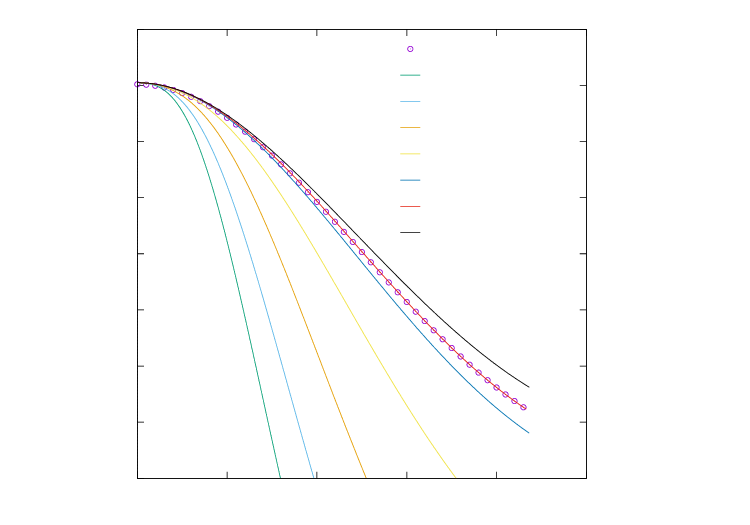}
}
\end{minipage}
\begin{minipage}[t]{0.49\columnwidth}
\centering
\hspace*{-5mm}
\scalebox{0.46}{
\input{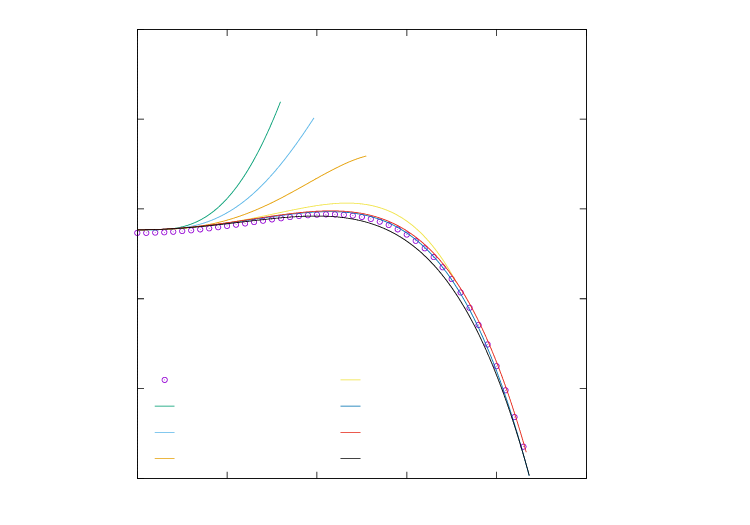}
}
\end{minipage}
\vspace*{3mm}

\begin{minipage}[t]{0.49\columnwidth}
\centering
\hspace*{-5mm}
\scalebox{0.46}{
\input{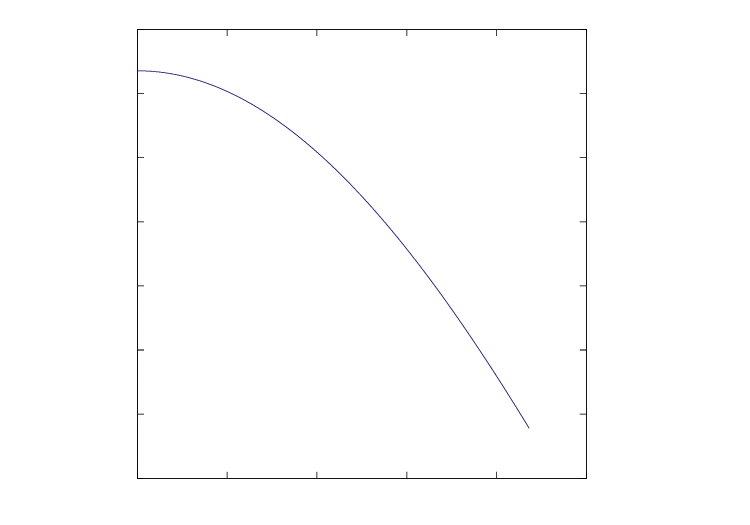}
}
\end{minipage}
\begin{minipage}[t]{0.49\columnwidth}
\centering
\hspace*{-5mm}
\scalebox{0.46}{
\input{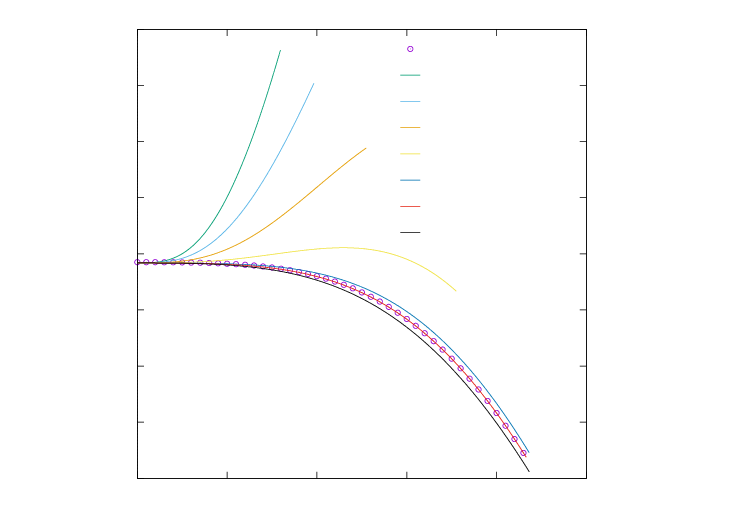}
}
\end{minipage}
\vspace*{1mm}
\caption{Gait descriptors with respect to $\beta$ for seven values of $\kappa$ where $T_{\rm set} = 0.7$ [s] of nonlinear and CLRed models}
\label{fig4.02}
\vspace*{-4mm}
\end{figure}

Figure \ref{fig4.02} shows the results of analyzing gait descriptors with respect to $\beta$ of the nonlinear model and CLRed models by setting $\kappa$ to seven different values. Here, (a) is the steady step period, $T_{\infty}$, (b) the steady angular velocity immediately before impact, $\dth_{1\infty}^-$, (c) the step length, and (d) the walking speed. All system parameters other than $\beta$ were set to the same values as those in Table \ref{table01}. Although the graphs in each figure have $\beta$ intervals set very finely at 0.001 [rad], by utilizing high-speed calculations with the CLRed model, computations for all $\beta$ ranges at a single $\kappa$ could be completed within a few minutes. The robot was set to an appropriate initial state and started walking; data from 20 steps after the 1000-th step was saved, and the average value was plotted. Calculation for a single point was completed in near-instantaneous time.

From the results of the nonlinear model, it can be seen that as $\beta$ increases, the step period and step length monotonically shorten, whereas the walking speed monotonically increases. Results from Figs. \ref{fig4.02}(a) and (d) show that the case with $\kappa = -0.5$ achieves a very high level of approximation accuracy for the values of steady step period and walking speed. From (b), however, it appears that the value of $\dth_{1{\infty}}^-$ is approximated with higher precision for $\kappa = -0.4$. Regarding the step length in (c), since it is uniquely determined by the values of $\alpha$ and $\beta$ in all cases, it was plotted using the theoretical value as a reference. The walking speed was calculated as the step length divided by the step period. Although both step length and step period decrease monotonically with increasing $\beta$, the fact that walking speed also decreases monotonically indicates that the reduction in step length has a greater influence.


\subsection{Descending Small Step}

\begin{figure}[!t]
\vspace*{0.5mm}
\centering
\footnotesize
\scalebox{0.65}{
\input{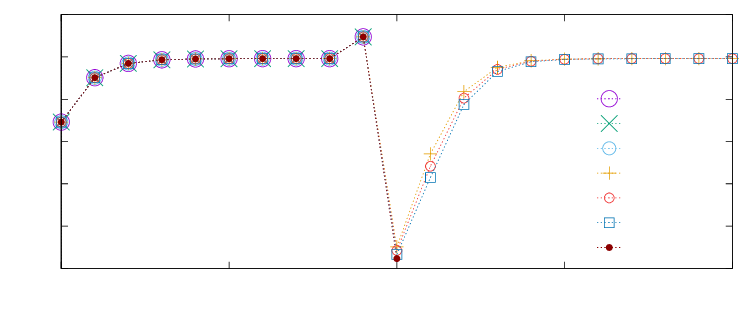}
}\\
\vspace*{-1mm}
(a) Step period \\
\scalebox{0.65}{
\input{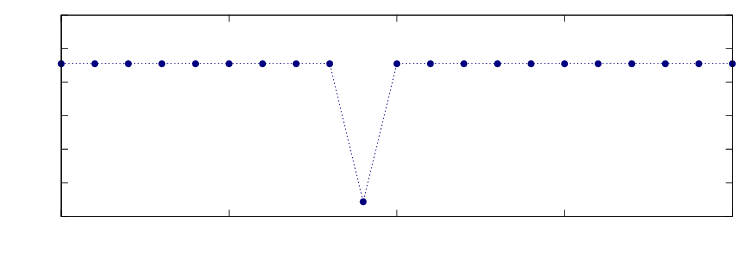}
}\\
\vspace*{-1mm}
(b) Step length\\
\vspace*{-2mm}
\caption{Evolution of gait descriptors during generated gaits involving descending 2.0 [cm] step for seven values of $T'_{\rm set}$ where $\beta = 0.7$ [rad]}
\label{fig4.03}
\vspace*{-3mm}
\end{figure}

This subsection considers the following walking situation. The robot begins walking from the target impact posture with an appropriate initial angular velocity, walks on a flat surface for a while, descends a 2.0 [cm] step at the 10-th impact, and then walks on a flat surface again. As a result, the 9-th step period becomes slightly longer, and the step length (horizontal distance traveled by the stance foot) of the 9-th step becomes slightly shorter, as shown in Figure \ref{fig4.03}. First, we investigated walkability using the CLRed model with $T_{\rm set} = 0.7$ [s], but forward motion accelerated immediately after descending the step, preventing control completion before the 11-th impact. Therefore, as shown in Fig. \ref{fig4.03}(a), data for step periods beyond the 10-th step was not saved. As a solution to this problem, we propose modifying $T_{\rm set}$ to be shorter $T'_{\rm set}$ only for the 10-th step and analyze its effect below.

This time, $\beta$ was set to $0.7$ [rad] and $\kappa = -0.5$, while all other system parameters were set to the same values as in Table \ref{table01} again. For all $T'_{\rm set}$ values, the robot performs exactly the same motion up until the 10-th impact, but starting from the 10-th step motion, it generates different movements for each case. From Fig. \ref{fig4.02}(a), it can be seen that when $T'_{\rm set}$ is set to three values, 0.55, 0.50 and 0.45 [s], the robot successfully descends the step and returns to steady walking. Fig. \ref{fig4.03}(b) plots the step length calculated in these successful cases. When $T'_{\rm set}$ is set to the three larger values, the absence of the 10-th step value indicates that the 10-th step motion could not be generated correctly. This is because $T'_{\rm set}$ was too long, preventing the control from completing before the 11-th impact, as described above. When $T'_{\rm set}$ is set to 0.40 [s], motion generation succeeded up to the 10-th step, but failed for the 11-th step. This occurred because the target settling time for the 10-th step motion was set too short, causing the forward motion to accelerate significantly. When resetting $T'_{\rm set}$ to $T_{\rm set}$ at the start of the 11-th step motion, control could not be completed before the 12-th impact.

In the gait generation considered in this subsection, however, $T_{\rm set}$ was changed to $T'_{\rm set}$ at the 10-th impact. Therefore, it is important to note that at this specific moment, not only $\bEta_{2i}$ but also $\bEta_1$ and $\bEta_3$ require recalculation based on the value of $T'_{\rm set}$. If control parameters such as $T_{\rm set}$, $\alpha$ and $\beta$ are adjusted every step to generate more robust walking modtion, $\bEta_1$ and $\bEta_3$ in Eq. (\ref{eq3.09}) must also be recalculated at each collision as $\bEta_{1i}$ and $\bEta_{3i}$, similar to $\bEta_{2i}$.

\section{CONCLUSION AND FUTURE WORK}

This paper discussed the mathematical modeling, dimensionality reduction, linearization, control system design, and motion analysis of the planar 6-DOF biped with all frames balanced around the hip joint. By utilizing the proposed CLRed model, it has become possible to investigate changes in gait characteristics corresponding to the stance-knee angle, $\beta$, as well as the feasibility of descending a small step, in a very short time. Furthermore, it was demonstrated that by appropriately setting the expansion point for linearization of the gravity term, a highly accurate CLRed model can be obtained regardless of the value of $\beta$. It is anticipated that this advantage unique to AL3 robots \cite{Kiefer,Spong,Agrawal,Agrawal2} will enable instantaneous walkability determination under various conditions. 

The biped robot model discussed in this paper does not exhibit the natural swinging motion or knee flexion/extension movements of the swing leg seen in passive dynamic walkers \cite{McGeer,McGeer2}. The authors believe that, however, this balanced body structure may reduce the risk of falls and facilitate the generation of highly stable walking motion. As reported in the previous section, it offers advantages in high-precision approximate linearization, and we anticipate it possesses various other potential benefits.

Section IV-C analyzed the walkability when descending a small step. It was revealed that despite the step height being very small compared to leg length, continued walking becomes impossible, and that adjusting only the target settling time could not generate sufficiently adaptive walking motion. In the future, we aim to develop a control strategy for generating more robust walking motion designed for challenging conditions, such as traversing stepping stones \cite{AIM2015,Li,Jenelten}. This strategy will incorporate adjustments not only to the target settling time but also to knee and hip joint angles, starting before approaching the step, based on visual perception of the ground surface conditions.


\end{document}